\documentclass[11pt]{article}

\usepackage[preprint]{acl}

\usepackage{times}
\usepackage{latexsym}

\usepackage[T1]{fontenc}

\usepackage[utf8]{inputenc}

\usepackage{microtype}
\usepackage{inconsolata}

\usepackage{graphicx}
\usepackage{booktabs}
\usepackage{multirow}
\usepackage{pdfpages}
\usepackage{float}
\usepackage{afterpage}
\usepackage{svg}
\usepackage{xcolor}
\usepackage{array}
\usepackage{longtable}
\usepackage{enumitem}

\title{StoryLens: Preference-Aligned Story Rewriting via Context-Aware Narrative Enrichment}

\author{
  \textbf{Hanwen Cui\thanks{Equal contribution.}\textsuperscript{1}},
  \textbf{Yuting Mei\footnotemark[1]\textsuperscript{2}},
  \textbf{Yuhang Fu\textsuperscript{1}},
  \textbf{Dingyi Yang\textsuperscript{3}},
  \textbf{Qin Jin\thanks{Corresponding author.}\textsuperscript{2}}
  \\
  \textsuperscript{1}Beijing University of Posts and Telecommunications
  \\
  \textsuperscript{2}AIM3 Lab, Renmin University of China
  \quad
  \textsuperscript{3}Nanyang Technological University
  \\
  \texttt{cuihanwen2023@bupt.edu.cn}
  \quad
  \texttt{\{meiyuting1004,qjin\}@ruc.edu.cn}
}

\begin{document}


\maketitle

\begin{abstract}
Story rewriting aims to adapt existing narratives to diverse reader preferences while preserving plot consistency and narrative coherence. Unlike conventional work on style transfer, we argue that effective story rewriting demands context-aware narrative enrichment beyond surface-level stylistic adaptation. 
Our pilot human study shows that style adaptation alone provides only marginal gains in reader satisfaction ($2.3\%$), while context-enhanced rewriting substantially improves user preference alignment ($24.5\%$).
Motivated by this, we introduce \textsc{StoryLensBench}, a large-scale benchmark for preference-aligned story rewriting, comprising structured story books, multi-dimensional reader preference profiles, and ranked context-aware rewritten stories.
Building on this benchmark, we propose \textsc{StoryLensEval}, a reward model for estimating reader satisfaction over rewritten stories, and \textsc{StoryLensWriter}, a two-stage rewriting model combining supervised fine-tuning with GRPO-based reinforcement learning. We further establish a comprehensive evaluation framework covering fidelity, coherence, and reader satisfaction. Experimental results demonstrate that \textsc{StoryLensWriter} consistently outperforms strong generation and personalization baselines, highlighting the importance of context-aware narrative enrichment for personalized story rewriting. \footnote{Code, data and model checkpoints will be released upon acceptance.}

\begin{figure}[t]
    \centering
    \includegraphics[
        width=\columnwidth,
        trim=49mm 2mm 7mm 0mm,
        clip
    ]{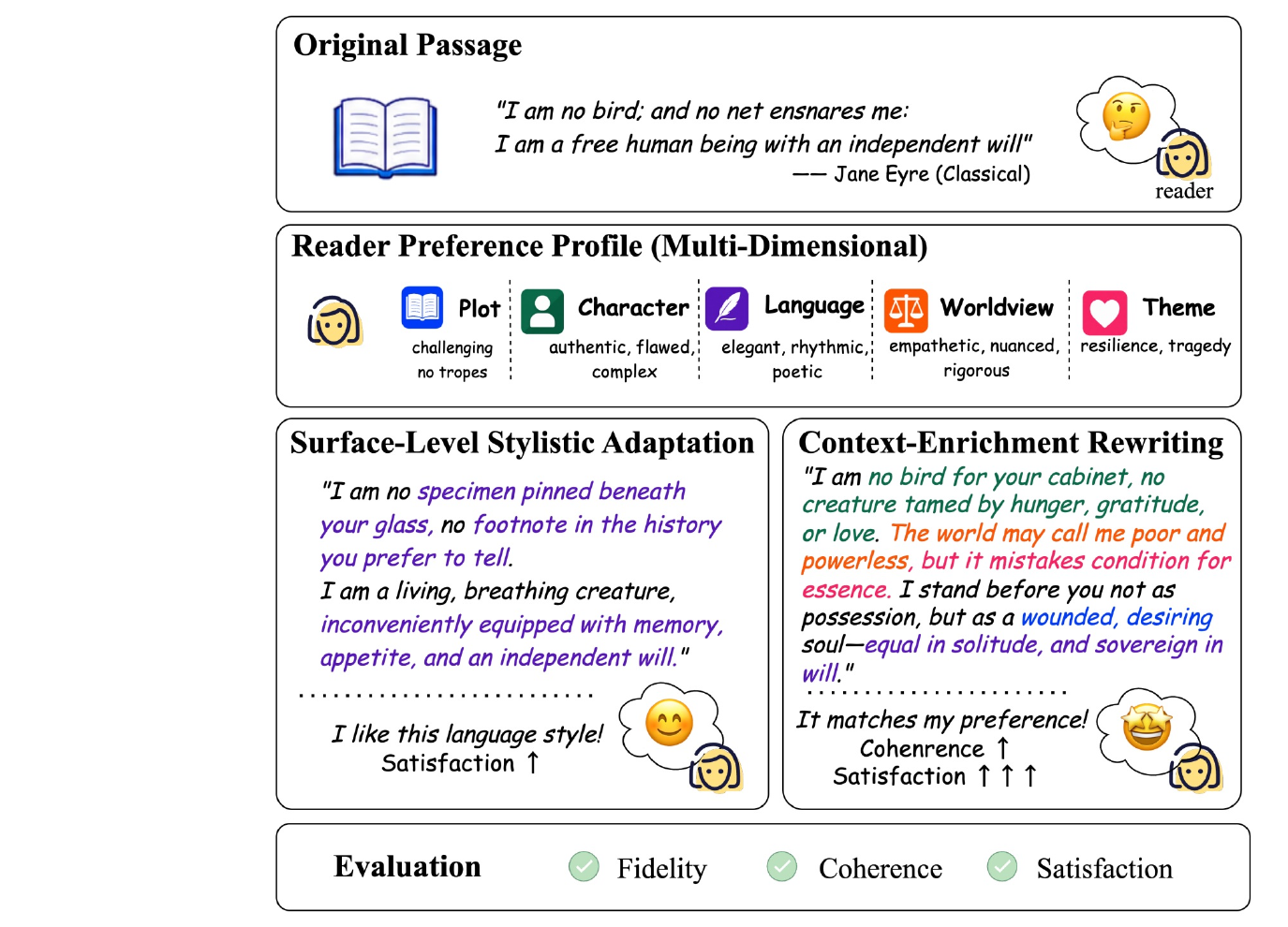}
    \caption{
    Overview of our story rewriting task. Given an original passage and a multi-dimensional reader preference profile, the model generates a rewritten story aligned with the target preferences while preserving the core narrative structure. 
    Unlike conventional style adaptation, our framework incorporates \emph{context-aware narrative enrichment}, which is shown through pilot studies to substantially improve reader satisfaction. The rewritten stories are evaluated along three dimensions: fidelity, coherence, and satisfaction.
    }
    \label{fig:intro_overview}
\end{figure}
%

\end{abstract}

\section{Introduction}

Story rewriting aims to adapt existing narratives to new audiences, communicative goals, and reader preferences while preserving the core plot, character consistency, and narrative coherence of the original story. 
Such capability is increasingly important for applications including personalized reading assistants, interactive storytelling, literary simplification, game narrative adaptation, and culturally localized content generation \citep{nasir2024word2world,huynh2025genai,singh2024translating}. 
Compared with open-ended story generation, story rewriting imposes a unique constraint: models must simultaneously preserve existing narrative structures while flexibly modifying the story according to user-specific preferences.

Despite recent advances in story generation \citep{fan2018hierarchical,yao2019plan,yang2022re3,yang2022doc,bae2024collective,park2025character}, personalized preference alignment remains largely underexplored. Existing controllable generation and style transfer methods mainly address sentence-level stylistic modification, such as sentiment, formality, or writing style adaptation \citep{liu2022semi,lai2024style,liu2025style}.
However, effective story rewriting for preference alignment goes well beyond surface-level stylistic transformation. To satisfy diverse reader preferences, models must perform \emph{context-aware narrative enrichment}, including expanding scene descriptions, refining emotional expression, enhancing character interactions, and elaborating narrative details, while maintaining global story consistency.
This makes story rewriting fundamentally different from conventional text editing or style transfer tasks.

We hypothesize that stylistic adaptation alone is insufficient for high-quality story rewriting. 
Instead, reader satisfaction depends heavily on whether the rewritten story can provide contextually appropriate narrative enrichment beyond simple lexical or stylistic changes. 
To verify this hypothesis, we conduct a pilot human study involving $10$ participants across $8$ books. 
Participants evaluate original stories, style-adapted rewrites, and context-enhanced rewrites on three dimensions: satisfaction, fidelity, and coherence. 
The results reveal that style adaptation alone provides only marginal improvement in reader satisfaction ($3.43 \rightarrow 3.51$), whereas context-enhanced rewriting raises satisfaction to $4.27$, with only minor reductions in fidelity and coherence.
These findings suggest that effective story rewriting requires modeling user preference and preference-aligned contextual enrichment.

To enable systematic study of this problem, we introduce \textbf{\textsc{StoryLensBench}}, a large-scale benchmark for preference-aligned story rewriting. 
The benchmark consists of chapter-level source stories, structured narrative contexts, multi-dimensional reader preference profiles, and ranked rewritten pairs. 
Unlike prior synthetic persona datasets \citep{jandaghi2024faithful,ge2024scaling}, \textsc{StoryLensBench} leverages real reader review histories collected from Goodreads\footnote{\url{https://www.goodreads.com/}}, enabling more realistic and context-aware preference modeling for long-form narratives. 
Our benchmark supports the study of personalized rewriting under realistic constraints where models must balance narrative fidelity, coherence, and user satisfaction simultaneously.

Building upon this benchmark, we further propose a rewriting framework consisting of two components: \textbf{\textsc{StoryLensEval}} and \textbf{\textsc{StoryLensWriter}}. 
\textsc{StoryLensEval} is a reward model designed to estimate reader satisfaction over rewritten stories conditioned on user preference profiles. 
Instead of relying on absolute quality scoring, we formulate evaluation as pairwise preference ranking to better capture subjective user preferences in narrative rewriting. 
Based on this evaluator, we develop \textsc{StoryLensWriter}, a preference-aligned rewriting model optimized through supervised fine-tuning followed by reinforcement learning with GRPO \citep{shao2024deepseekmath}. 
The model learns to generate rewrites that not only preserve narrative consistency but also perform contextually appropriate narrative enrichment aligned with reader preferences.

We further establish a comprehensive evaluation protocol for story rewriting across three key dimensions: \emph{fidelity}, \emph{coherence}, and \emph{satisfaction}. 
Fidelity measures whether rewritten stories preserve the core narrative content of the original story. 
Coherence evaluates narrative consistency and writing fluency across long contexts. 
Satisfaction measures how well rewrites align with user preferences through pairwise comparison-based evaluation.

In summary, our contributions are threefold:

\begin{itemize}[leftmargin=*, itemsep=1pt, topsep=1pt, parsep=0pt]

\item We formulate personalized story rewriting as a context-aware narrative adaptation task that extends beyond conventional style transfer and requires creative narrative enrichment for improving reader satisfaction.

\item We introduce \textsc{StoryLensBench}, the first large-scale benchmark for preference-aligned long-form story rewriting with real reader preference profiles and ranked rewritten stories.

\item We propose \textsc{StoryLensEval} and \textsc{StoryLensWriter}, enabling preference-aware evaluation and context-enhanced rewriting generation through supervised fine-tuning and reinforcement learning.

\end{itemize}

\section{Related Work}

\begin{figure*}[t]
    \centering
    \includegraphics[
        width=\textwidth,
        trim=12mm 50mm 12mm 50mm,
        clip
    ]{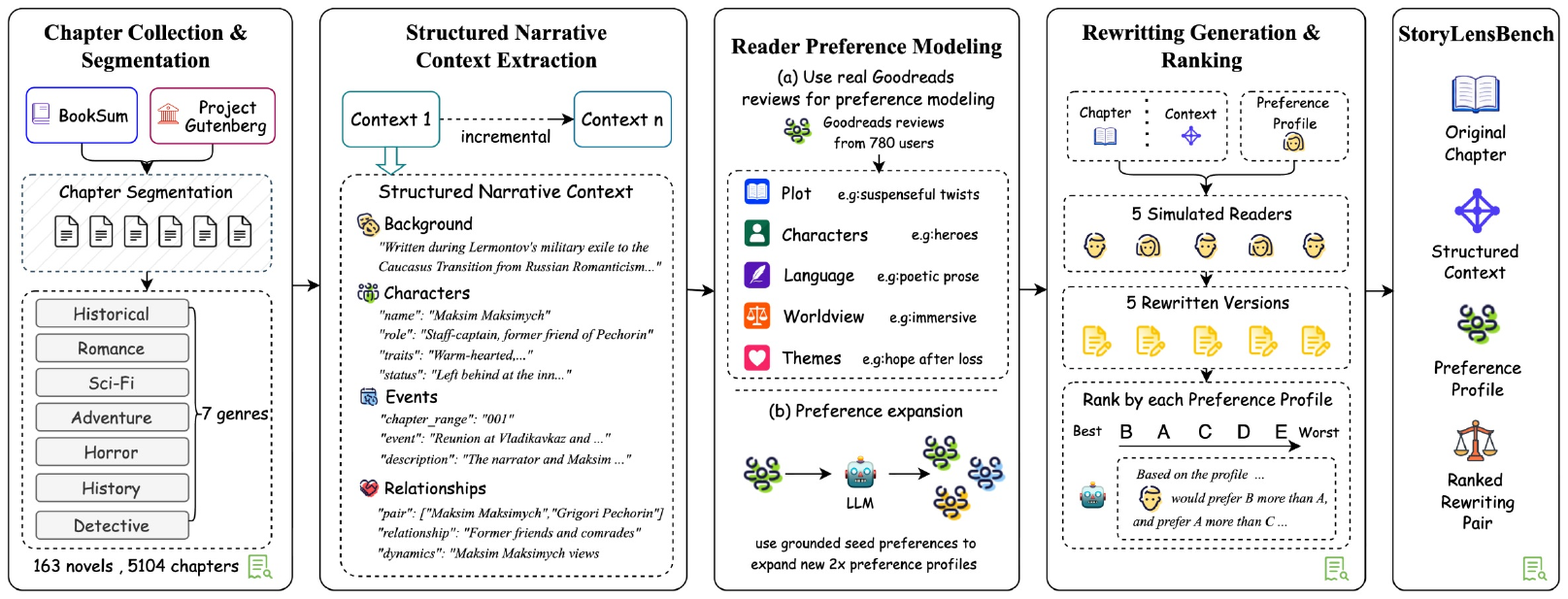}
    \caption{
    Overview of the dataset construction process.
    The green document icon in the bottom right corner indicates that this step has been manually checked.
    }
    \label{fig:datasets}
\end{figure*}

\noindent \textbf{Story Generation.}
Recent Large Language Models (LLMs) show strong abilities in creative writing and story generation~\citep{zhao2023survey}, with prior work exploring incremental generation~\citep{yang2022re3}, hierarchical planning~\citep{yang2022doc, wang2024dome}, and multi-agent collaboration~\citep{xia2025storywriter, yu2025multi}.
However, these methods mostly generate stories for general audiences, whereas our work focuses on adapting existing stories to reader-specific preferences while preserving narrative consistency.

\noindent \textbf{Personalization of LLMs and Evaluation.}
Personalization of LLMs adapts outputs to user-specific contexts, profiles, or feedback. 
Existing approaches include prompt engineering, RAG-based methods, representation learning methods, and RLHF-based methods~\citep{zhang2024personalization}. 
Prompting engineering directly encodes personas or profiles in instructions for LLMs~\citep{jiang2023evaluating,lyu2024llm}.
RAG-based methods further retrieve user histories or external memories~\citep{gao2023retrieval,salemi2024lamp} into instructions to augment text generation. 
Representation learning methods compress user information into learned embeddings~\citep{liu2025persona_plug,ning2025user} for LLMs.
The aforementioned methods confine personalization modeling to the LLM's input stage, limited by the capabilities of the model itself.
RLHF-based methods optimize LLMs' own capacity with preference feedback~\citep{ouyang2022training,li2024personalized,lee2024aligning}. However, these methods target short dialogue and encode preferences either as latent embeddings or textual descriptions, requiring substantial data to generalize to new preference types.
As it is costly to annotate large amounts of preference feedback data for long-form stories, we model real reader preferences as multi-dimensional signals and learn a preference reward model to train the story rewriting model.

Evaluation is also non-trivial for story rewriting. 
Existing metrics of story generation often focus on either fidelity or style: fidelity-oriented methods evaluate factual consistency or plot preservation~\citep{wang2020asking,chang2023booookscore,kim2024fables}, while style-oriented methods measure linguistic similarity, persona adherence, or personalized preference alignment~\citep{wegmann2021does,lin2023llm,wang-etal-2024-learning-personalized,salemi2025expert}. 
However, these standard metrics are often too rigid for rewriting, as they penalize reasonable contextual enrichment or stylistic re-expression. 
Unlike these metrics, \textsc{StoryLensEval} estimates reader satisfaction given a preference profile, rather than measuring surface-level style similarity or aspect-level alignment.

\section{\textbf{\textsc{StoryLensBench}}}

\subsection{Benchmark Construction}

\textsc{StoryLensBench} consists of four components: original novel chapters, structured contexts, reader preference profiles, and ranked rewriting pairs.
We construct our data via a unified four-step pipeline, as illustrated in Figure~\ref{fig:datasets}:
First, we collect and segment novels into chapter-level source texts. 
Second, we extract structured narrative contexts from the preceding storyline for each chapter to serve as critical global constraints. 
Third, we reverse-engineer and model fine-grained reader preference profiles based on real-world book reviews. 
Finally, we synthesize various preference-tailored rewrite candidates for each chapter and subsequently rank them according to their alignment with the target preferences.
We next explicate each step in turn.

\noindent\textbf{Chapter Collection \& Segmentation}\quad
We collect source novels from BookSum~\citep{kryscinski2022booksum} and Project Gutenberg. 
While BookSum inherently provides chapter-partitioned narratives, long-form novels from Project Gutenberg  are segmented into chapters based on structural markers. 
After deduplication, content filtering, and genre balancing, we secure a clean corpus of 163 chapter-segmented novels.
The final collection spans seven diverse genres: Historical, Romance, Sci-Fi, Adventure, Horror, Humor, and Detective.

\noindent\textbf{Structured Narrative Context Extraction}\quad
To support coherent long-form rewriting, we construct a structured narrative context for each chapter based on its preceding storyline. 
This context encapsulates the novel's writing background, character information, major events, and character relations. 
Crucially, the extraction of structured context shifts the personalized rewriting task from simple surface-level text style transfer to deep, context-aware narrative enrichment. 
By injecting these global continuity signals, we empower the model to actively enrich chapter details and introduce creative expansions while remaining strictly anchored to the established plot logic~\citep{yang2025matters}.

\noindent\textbf{Reader Preference Modeling}\quad
Rather than relying on artificial or generic descriptions, we model authentic user behavior by crawling real-world reviews on Goodreads from 780 active readers who engaged with the corresponding source books.
We leverage Gemini 3.1 Pro~\citep{google2026gemini31pro} to distill these open-ended reviews into structured reading preferences. 
To capture the full complexity of reader intent beyond mere linguistic style, each preference profile is organized into a fine-grained five-dimensional taxonomy: \texttt{Plot}, \texttt{Characters}, \texttt{Language}, \texttt{Worldview}, and \texttt{Theme}.
We further synthesize $2\times$ additional profiles from seed profiles.
We manually check the profiles to ensure authenticity, plausibility, and diversity.

\noindent\textbf{Rewriting Generation \& Ranking}\quad
For each novel, we sample five distinct reader profiles and utilize Gemini 3.1 Pro to generate five variant rewrites for each chapter, augmenting the prompts with the extracted structured narrative context to guide the enrichment process. 
All rewrite candidates are then ranked according to their adherence to each targeted reader profile. 
Consecutive variants within the ranked sequence are subsequently transformed into pairwise preference comparisons to facilitate subsequent alignment training.

To validate data quality, we conduct human verification on randomly sampled cases.
Specifically, each of the 10 annotators rates 10 sampled rewrites on fluency, fidelity, and preference alignment using a 0--3 Likert scale, and evaluates 10 ranking pairs for ranking reasonableness. As shown in Table~\ref{tab:human_verification}, the relative rankings achieve a high average inter-annotator agreement of 81\%, confirming the rigorous quality of our benchmark.

\subsection{Data Analysis}

An empirical analysis of \textsc{StoryLensBench} (Figures~\ref{fig:goodreads_demographics}, \ref{fig:book_overview}, and \ref{fig:top_keywords_by_field}) reveals three core characteristics:
i) the \textbf{reader distribution} spans a heterogeneous spectrum of ages and professional backgrounds from authentic reviews, capturing realistic user personas rather than artificial constructs. 
ii) the \textbf{book distribution} encompasses balanced literary genres and maintains a distinct long-tailed text-length distribution, establishing a challenging testbed that necessitates  narrative enrichment rather than surface style transfer. 
iii) the \textbf{preference vocabulary} follows a salient long-tail pattern, demonstrating that the corpus successfully encapsulates both global reading trends and fine-grained, highly personalized reader nuances.





\section{Methodology}
\begin{figure*}[t]
    \centering
    \includegraphics[
        width=\textwidth,
        trim=3mm 74mm 2mm 74mm,
        clip
    ]{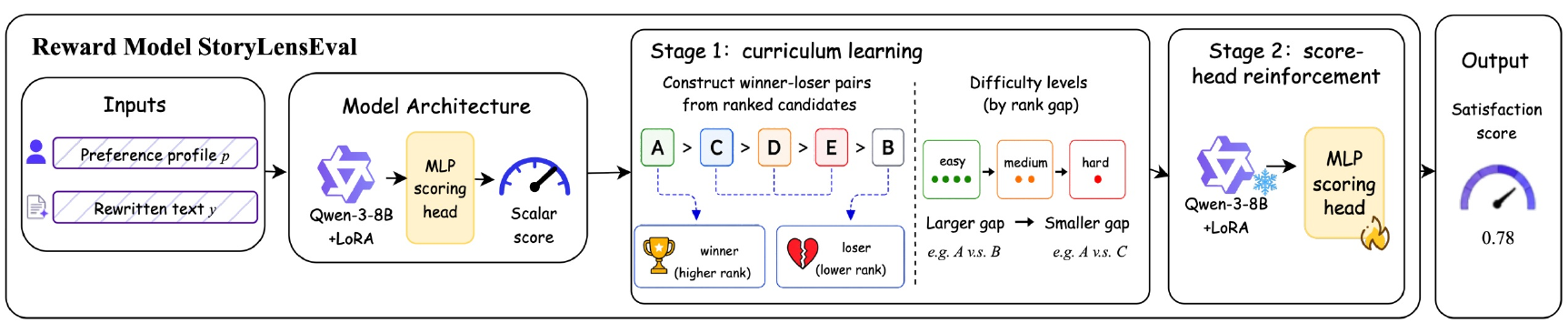}
    \caption{
    Overview of the training process for \textbf{\textsc{StoryLensEval}}.
    }
    \label{fig:train1}
\end{figure*}
\subsection{Task Formulation}
We formally define the preference-aligned story rewriting task as follows.
Given an original story segment $x$, its structured narrative context of preceding contents $c$, and a target user preference profile $p$, preference-aligned story rewriting aims to generate an optimal rewrite $y^{*}$:
\[
y^{*} = \arg\max_{y} P_{\theta}(y \mid x, c, p),
\]
where the rewrite should preserve the main plot line of $x$, remain consistent with the structured context $c$, and satisfy the target reader's preferences $p$. 
Story rewriting entails two main challenges: preference alignment evaluation and context-aware text generation. 
To address these challenges, we first train a model that measures how well a rewrite matches a preference profile. Leveraging this as a reward signal, we further develop a rewriting model via a two-stage training pipeline.

\begin{figure*}[t]
    \centering
    \includegraphics[
        width=\textwidth,
        trim=2mm 70mm 2mm 74mm,
        clip
    ]{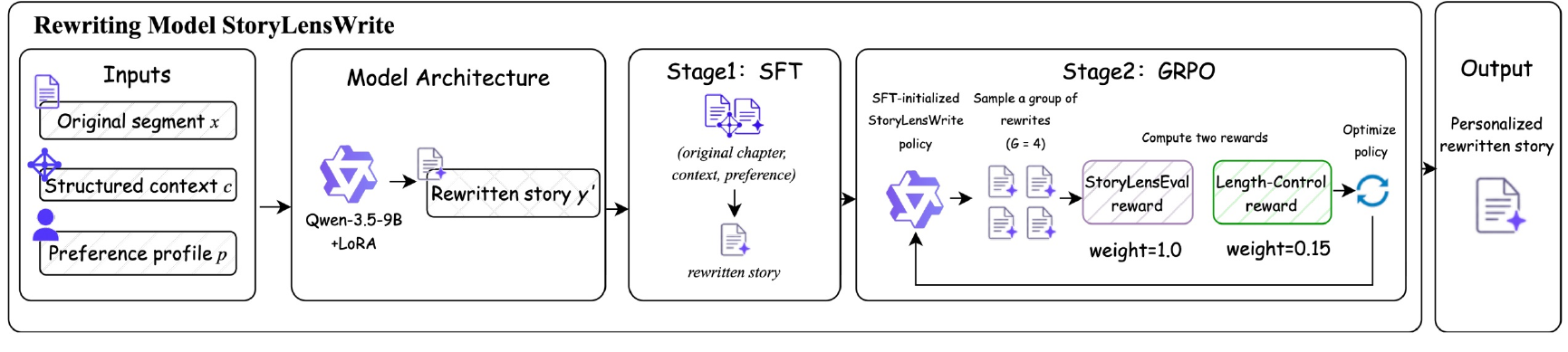}
    \caption{
    Overview of the training framework for \textbf{\textsc{StoryLensWriter}}.
    }
    \label{fig:train2}

\end{figure*}
\subsection{\textsc{StoryLensEval}}

We first train \textsc{StoryLensEval}, which measures the degree to which a rewritten text matches the target reader's preferences on our preference-ranking dataset \textsc{StoryLensBench}. 

Our training data is formatted as ranked rewrite pairs conditioned on specific reader preferences.
As illustrated in Figure~\ref{fig:train1}, each instance in \textsc{StoryLensBench} contains multiple candidate rewrites ranked by their alignment with a user preference profile. 
We transform these rankings into pairwise comparisons, treating the higher-ranked rewrite as the winner and the lower-ranked one as the loser. 
Based on the ranking gap between two candidates, these pairs are categorized into easy (larger gap), medium, and hard (smaller gap) levels.

\textsc{StoryLensEval} adopts a Qwen3-8B backbone~\citep{yang2025qwen3} with an MLP scoring head, and is trained with LoRA\citep{hu2022lora} for parameter-efficient adaptation.
Given a user preference profile $p$ and a rewritten text $y$, the model outputs a scalar satisfaction logit:
\begin{equation}
s = f_{\theta}(p, y),
\end{equation}
where $f_{\theta}$ denotes the trained satisfaction scoring model.

The model is optimized with a pairwise ranking objective:
\begin{equation}
\mathcal{L}_{sat} = \mathrm{softplus}(-(s_w - s_l)),
\end{equation}
where $s_w$ and $s_l$ denote the scores of the winner and loser, respectively. 
This objective encourages the model to assign higher scores to rewrites that better satisfy the user's preferences. 

We train the model in two stages: an easy-to-hard curriculum learning phase, followed by scoring head refinement to better distinguish hard comparison pairs and improve final scalar scoring precision.

\subsection{\textbf{\textsc{StoryLensWriter}}}

We adopt a two-stage training strategy to build our rewriting model.
Specifically, we use Qwen3.5-9B~\citep{qwen2026qwen35_9b} as the backbone model and progressively adapt it to the personalized story rewriting task.
The first stage performs supervised fine-tuning (SFT) to teach the model the task format and basic rewriting behavior, while the second stage further optimizes the model with reinforcement learning to improve preference alignment and controllability.

\subsubsection{Stage 1: Supervised Fine-Tuning (SFT)}

In the first stage, we fine-tune Qwen3.5-9B with LoRA adapters using the paired rewriting data from {\textsc{StoryLensBench}}.
This stage aims to teach the model the basic input-output format of the task and provide it with an initial ability to perform preference-aware story rewriting.

Given an original segment $x$, a structured context $c$, and a user preference profile $p$, the model is trained to generate the target rewrite $y$.
We format the input as an instruction-following prompt that explicitly requires the model to preserve the main plot, maintain consistency with the context, and adapt the rewritten text to the user's preferences.
The SFT objective is the standard autoregressive language modeling loss:
\begin{equation}
\mathcal{L}_{\mathrm{SFT}}
=
-\sum_{t=1}^{|y|}
\log P_{\theta}(y_t \mid y_{<t}, x, c, p).
\end{equation}

\subsubsection{Stage 2: Reinforcement Learning with GRPO}

Although SFT enables the model to imitate reference rewrites, it does not directly optimize the model toward preference alignment or task-specific quality criteria.
Therefore, in the second stage, we further optimize the SFT model using Group Relative Policy Optimization (GRPO)\citep{shao2024deepseekmath}. 
GRPO optimizes the policy by increasing the likelihood of candidates with relatively higher rewards, encouraging the model to better balance user preference alignment and practical rewriting constraints.

To construct the training set, we select approximately 5\% of the SFT data and convert them into prompts containing the original story segment, structured narrative context, and user preference profile.
For each prompt, the policy model samples a group of candidate rewrites.
We use {\textsc{StoryLensEval}}, the preference scoring model introduced in the previous subsection, as the reward model.
This reward model assigns a scalar score $R_{\mathrm{rm}}$ to each generated rewrite, measuring how well it satisfies the target user's preferences.

To avoid degenerate length changes, we additionally introduce a length-control reward.
Specifically, we encourage the rewritten text to remain within $0.7$--$1.3$ times the length of the original segment.
When the generated rewrite falls outside this range, we apply a Huber-style penalty $R_{\mathrm{length}}$ based on the deviation from the valid interval\citep{huber1992robust}.

The final reward is defined as:
\begin{equation}
R = R_{\mathrm{rm}} + 0.15 R_{\mathrm{length}}.
\end{equation}


\begin{table*}[t]
\centering
\small
\resizebox{\textwidth}{!}{
\begin{tabular}{lccccc}
\toprule
\multicolumn{1}{c}{\multirow{2}{*}{\textbf{Model}}}
& \multicolumn{4}{c}{\textbf{Narrative Quality}}
& \multicolumn{1}{c}{\textbf{Preference Alignment}} \\
\cmidrule(lr){2-5} \cmidrule(lr){6-6}
& \textbf{Local Fid. $\uparrow$}
& \textbf{Global Fid. $\uparrow$}
& \textbf{Coh. $\uparrow$}
& \textbf{NovelCritique $\uparrow$}
& \textbf{PerSE $\uparrow$} \\
\midrule
Gemini-3.1-Pro
& 93.68
& 97.54
& 92.65
& 83.80
& \underline{7.69} \\

GPT-5.2
& 93.61
& \underline{98.97}
& \textbf{97.01}
& \textbf{88.08}
& 7.59 \\

DeepSeek-V3
& 93.50
& 98.56
& 92.91
& 85.86
& 7.67 \\

Doubao-Seed-2.0-Lite
& 93.86
& \underline{98.97}
& 93.40
& 86.54
& 7.62 \\

Qwen-VL-Max
& \underline{94.45}
& 98.44
& \underline{93.82}
& 85.53
& \underline{7.69} \\

\textbf{\textsc{StoryLensWriter}}
& \textbf{94.62}
& \textbf{98.98}
& 93.11
& \underline{86.93}
& \textbf{7.78} \\
\bottomrule
\end{tabular}
}
\caption{
Comparison with strong LLMs. Metrics are grouped into narrative quality and preference alignment.
}
\label{tab:strong_llm_results}
\end{table*}

\begin{table*}[t]
\centering
\small
\resizebox{\textwidth}{!}{
\begin{tabular}{lccccccc}
\toprule
\multicolumn{1}{c}{\multirow{2}{*}{\textbf{Model / Method}}}
& \multicolumn{4}{c}{\textbf{Narrative Quality}}
& \multicolumn{3}{c}{\textbf{Preference Alignment}} \\
\cmidrule(lr){2-5} \cmidrule(lr){6-8}
& \textbf{Local Fid. $\uparrow$}
& \textbf{Global Fid. $\uparrow$}
& \textbf{Coh. $\uparrow$}
& \textbf{NovelCritique $\uparrow$}
& \textbf{Sat. Win $\uparrow$}
& \textbf{Sat. Top-1 $\uparrow$}
& \textbf{PerSE $\uparrow$} \\
\midrule
Llama-3.1-8B
& 92.54
& 96.96
& 80.53
& 78.55
& 0.09
& 0.00
& 7.52 \\

Qwen3.5-9B
& 94.12
& {98.64}
& 90.72
& 85.78
& 0.52
& 0.03
& \underline{7.68} \\

Hybrid RAG (Llama-3.1-8B)
& 92.29
& 93.26
& 79.64
& 68.00
& 0.21
& 0.02
& 7.37 \\

Hybrid RAG (Qwen3.5-9B)
& 93.97
& 96.26
& 87.54
& 81.43
& 0.64
& 0.18
& 7.50 \\

PPlug
& \underline{94.35}
& \textbf{99.71}
& 80.99
& \underline{86.14}
& \underline{0.70}
& \underline{0.30}
& 7.65 \\

\textbf{\textsc{StoryLensWriter}}
& \textbf{94.62}
& \underline{98.98}
& \textbf{93.11}
& \textbf{86.93}
& \textbf{0.84}
& \textbf{0.47}
& \textbf{7.78} \\
\bottomrule
\end{tabular}
}
\caption{
Comparison with comparably-sized open-source models and representative personalization methods.
}
\label{tab:personalization_results}
\end{table*}

\subsection{Evaluation Framework}

To comprehensively assess the quality of rewriting, we establish a multi-dimensional evaluation framework across three core aspects: fidelity, coherence, and reader satisfaction.

\noindent\textbf{\textsc{Fidelity.}} 
Fidelity consists of two parts: local fidelity and global fidelity.

\textit{Local Fidelity }
Local fidelity uses only the original and rewritten segments as input, and evaluates whether the rewrite preserves the language, main plot, entities, and settings of the original passage. 

Inspired by ~\citep{li2025storyteller}, we calculate local fidelity based on Subject-Verb-Object (SVO) triplets and Narrative Entity Knowledge Graphs (NEKG).
We use GPT-4o-mini~\citep{openai2024gpt4o_mini} to extract SVOs and NEKGs from both the original and rewritten texts.
For SVOs, we use text-embedding-3-small\citep{openai2024textembedding3small} to find the best-matching pairs between the two versions. 
If the cosine similarity falls into a fuzzy range, i.e., 0.50--0.85, GPT-4o makes a binary arbitration on whether the item is faithfully preserved.
For NEKG, we use GPT-4o to determine whether each entry is preserved.

For each component $i \in \{\mathrm{SVO}, \mathrm{NEKG}\}$, we denote its pass rate as $S_i$, computed as the proportion of extracted items that pass the final judgment.
The final local fidelity score is calculated as:
\begin{equation}
F_{\mathrm{local}} = 0.9 \cdot S_{\mathrm{NEKG}} + 0.1 \cdot S_{\mathrm{SVO}}.
\end{equation}

\textit{Global Fidelity }
Global fidelity introduces the structured narrative context as an additional input to check whether the rewritten content remains consistent with the broader story setting. 
Following the conflict types in \citet{kim2024fables}, we use GPT-4o-mini to extract claims from the rewritten text.
Each claim is first compared against the original text. 
If it is not supported by the original text, it is further compared against the structured context. 
We categorize each claim into one of three labels: Entailment, Neutral, or Contradiction. 
Entailment indicates that the claim is fully supported, Neutral indicates a reasonable expansion, and Contradiction indicates a conflict with the original text or broader context.
The global fidelity score is defined as the proportion of claims labeled as Entailment or Neutral:
\begin{equation}
F_{\mathrm{global}} = 
\frac{N_{\mathrm{entail}} + N_{\mathrm{neutral}}}
{N_{\mathrm{total}}}.
\end{equation}

\noindent
\textbf{\textsc{Coherence.}}
Coherence evaluates the sentence-level fluency and logical consistency of the rewritten text.
Following the eight error types defined by \citet{chang2023booookscore}, we first segment each rewrite into sentences using punctuation delimiters, and then use GPT-4o-mini to detect potential coherence errors for each sentence.
The coherence score is computed as $C = N_{\mathrm{error\_free}} / N_{\mathrm{total}}$, where $N_{\mathrm{error\_free}}$ denotes the number of sentences without detected errors and $N_{\mathrm{total}}$ denotes the total number of sentences.

\noindent
\textbf{\textsc{Satisfaction.}}
Satisfaction measures the degree to which a rewritten text matches the target user's preferences.
We feed the user preference profile and rewritten text into \textsc{StoryLensEval}, and use the predicted scalar score as the satisfaction score:
\begin{equation}
S_{sat} = f_{\theta}(p, y).
\end{equation}

\label{sec:experiments}

\section{Experiments}

\subsection{Evaluation Dataset}

For evaluation, we construct a held-out test set consisting of newly collected stories and preference profiles. 
Specifically, we collect 29 additional novels from Project Gutenberg, covering the same seven story categories, with the number of novels kept as balanced as possible across categories. 
Following our preprocessing pipeline, we segment the novels into chapters, resulting in a total of 867 chapter-level samples for evaluation.
We also generate five new user preference profiles, deliberately maximizing their diversity across different dimensions.

\subsection{Evaluation Metrics}

We evaluate different rewriting methods from two perspectives: \textit{Narrative Quality} and \textit{Preference Alignment}. 
For \textit{Narrative Quality}, we use our proposed \textbf{fidelity} and \textbf{coherence} metrics.
We also use NovelCritique~\citep{yang2025matters} to assess overall long-story quality.

For \textit{Preference Alignment}, we use two complementary metrics: 
\textbf{PerSE} under the Per-MPST setting~\citep{wang-etal-2024-learning-personalized} 
and our \textit{satisfaction} evaluator \textbf{\textsc{StoryLensEval}}. 
PerSE serves as an external evaluation signal independent of the training objective of \textsc{StoryLensWriter}. 
Since \textsc{StoryLensEval} is trained as a task-specific relative evaluator, we report pairwise win rate and top-1 rate rather than absolute scores. 
To mitigate potential self-evaluation bias arising from our LLM-driven data construction, we utilize \textsc{StoryLensEval} primarily to benchmark open-source personalizable baselines.

We validate \textsc{StoryLensEval} against both human annotations and strong LLM judgments. Specifically, 40 randomly sampled story pairs are evaluated by three human annotators via majority voting, achieving a high inter-annotator correlation of 92.98\%.

\subsection{Baselines}
\label{subsec:baselines}

We compare \textsc{StoryLensWriter} with two categories of baselines, spanning strong general LLMs and representative personalization methods~\citep{zhang2024personalization}. 

\noindent\textbf{Strong LLMs}\quad
We directly prompt strong off-the-shelf LLMs with the same instruction, narrative context, and reader preference profile. Models include Gemini-3.1-Pro~\citep{google2026gemini31pro}, GPT-5.2~\citep{openai2025gpt5_2}, DeepSeek-V3~\citep{deepseek2024v3}, Doubao-Seed-2.0-Lite~\citep{volcengine2026doubao2}, and Qwen-VL-Max~\citep{bai2025qwen25vl}.

\noindent\textbf{Peer Open-source Models}\quad
We also compare with Llama-3.1-8B~\citep{meta2024llama3} and Qwen3.5-9B~\citep{qwen2026qwen35_9b}. 

\noindent\textbf{Hybrid RAG}\quad
We evaluate retrieval-based personalization by building a retrieval repository from the training set. For each test instance, we use hybrid retrieval with both dense and sparse matching to select preference-relevant examples, and then include them as few-shot prompting demonstrations. We apply hybrid RAG to both Llama-3.1-8B and Qwen3.5-9B.

\noindent\textbf{Representation-based Personalization}\quad
We also compare with Persona-Plug~\citep{liu2025persona_plug}, which encodes user information into a plug-and-play personalized embedding. We adapt it to our task and use the learned preference representation to guide story rewriting.

\begin{table}[t]
\centering
\small
\resizebox{0.98\columnwidth}{!}{
\begin{tabular}{lcc}
\toprule
\textbf{Evaluator}
& \textbf{Human Corr.}
& \textbf{Strong LLM Corr.} \\
\midrule
Qwen3-8B 
& 0.4211
& 0.4211 \\

Llama-3.1-8B
& 0.7368
& 0.7368 \\

\textbf{\textsc{StoryLensEval}}
& \textbf{0.8947}
& \textbf{0.9250} \\
\bottomrule
\end{tabular}
}
\caption{
Correlation with human annotations and strong LLM judgments.
}
\label{tab:eval_validation}
\end{table}

\subsection{Main Results}
\label{subsec:main_results}

\quad \textbf{\textit{{StoryLensWriter} improves preference alignment over prompted strong LLMs while preserving strong narrative quality. }}
Table~\ref{tab:strong_llm_results} shows that \textsc{StoryLensWriter} achieves the best PerSE score among the compared strong LLMs, indicating stronger alignment with target reader preferences. Our model obtains the highest local fidelity and global fidelity, suggesting that task-specific training on \textsc{StoryLensBench} helps the model ground preference-driven rewriting in the original passage and structured narrative context.

\textbf{\textit{Our proposed training framework is effective for both narrative quality and preference alignment. }}
Table~\ref{tab:personalization_results}  shows that \textsc{StoryLensWriter} outperforms comparably-sized open-source backbones, retrieval-based personalization, and representation-based personalization methods. 
This demonstrates that merely retrieving historical contexts or injecting persona representations is insufficient for rewriting. Instead, explicitly optimizing the model via our two-stage training paradigm genuinely instills a fine-grained understanding of multi-dimensional reader preferences.

\textbf{\textit{{StoryLensEval} provides a reliable relative preference signal for story rewriting. }}
As reported in Table~\ref{tab:eval_validation}, \textsc{StoryLensEval} achieves higher agreement with human annotations and strong LLM judgments than comparably-scale LLM judges. 
This suggests that preference-ranking training enables \textsc{StoryLensEval} to better capture reader-preference satisfaction, making it suitable as both an evaluator and a reward model for preference-aligned story rewriting.

\subsection{Ablation Study}

\label{subsec:ablation}
We conduct two ablation studies to analyze the effectiveness of our design choices. 
The first study focuses on \textsc{StoryLensEval}, including its model architecture and second-stage score-head reinforcement. 
The second study evaluates the RL stage of \textsc{StoryLensWriter} by comparing the full model with its SFT-only variant.

As shown in Table~\ref{tab:storylenseval_ablation}, replacing the single-tower design with a dual-tower architecture leads to a clear drop.
This suggests that jointly encoding the user preference profile and rewritten text is important for capturing fine-grained preference interactions. 
Removing the second training stage also decreases performance, especially on hard pairs, indicating that score-head reinforcement improves the discriminative ability of the model.

Table~\ref{tab:storylenswriter_ablation} shows the effect of the GRPO stage. 
Compared with the SFT-only variant, \textsc{StoryLensWriter} achieves better preference alignment after RL optimization. 
This indicates that using \textsc{StoryLensEval} as a reward signal guides the model toward preference-aligned rewriting.

\begin{table}[t]
\centering
\small
\resizebox{\columnwidth}{!}{
\begin{tabular}{lcccc}
\toprule
\textbf{Model}
& \textbf{Easy}
& \textbf{Medium}
& \textbf{Hard}
& \textbf{Overall} \\
\midrule
\textsc{StoryLensEval} w/ dual tower
& 83.26
& 74.27
& 65.36
& 73.40 \\

\textsc{StoryLensEval} w/o phase2
& 95.72
& 87.63
& 76.67
& 85.67 \\

\textbf{\textsc{StoryLensEval}}
& \textbf{95.85}
& \textbf{87.85}
& \textbf{76.91}
& \textbf{85.87} \\
\bottomrule
\end{tabular}
}
\caption{
Ablation study of \textsc{StoryLensEval}. 
}
\label{tab:storylenseval_ablation}
\end{table}

\begin{table}[t]
\centering
\small
\resizebox{\columnwidth}{!}{
\begin{tabular}{lcccc}
\toprule
\textbf{Model}
& \textbf{Fid.}
& \textbf{Coh.}
& \textbf{PerSE}
& \textbf{Win} \\
\midrule
\textsc{StoryLensWriter} w/o GRPO
& 94.04 / 97.51
& 91.10
& 7.59
& 20.44 \\

\textbf{\textsc{StoryLensWriter}}
& \textbf{94.62 / 98.98}
& \textbf{93.11}
& \textbf{7.78}
& \textbf{79.56} \\
\bottomrule
\end{tabular}
}
\caption{
Ablation study of \textsc{StoryLensWriter}.
}
\label{tab:storylenswriter_ablation}
\end{table}

\section{Conclusion}

In this paper, we study preference-aligned story rewriting, aiming to adapt literary narratives to diverse reader preferences while preserving narrative consistency. We make three contributions: (i) constructing \textsc{StoryLensBench}, a story rewriting benchmark with narrative contexts, reader preference profiles, and ranked rewrites; (ii) establishing an evaluation framework covering fidelity, coherence, and satisfaction, supported by \textsc{StoryLensEval}; and (iii) developing \textsc{StoryLensWriter}, a two-stage rewriting model trained with SFT and reinforcement learning. Experiments show that \textsc{StoryLensWriter} improves preference alignment over strong LLMs and personalization baselines while maintaining strong narrative quality. 

\section*{Limitations}
Our current experiments focus on English literary narratives and reader preference profiles derived from Goodreads. This setting provides a controlled testbed for preference-aligned story rewriting, but it may not fully reflect multilingual, cross-cultural, or highly contemporary reading preferences. Extending the benchmark to broader languages, cultures, and narrative domains is a natural direction for future work.

The source texts in our current benchmark mainly come from literary narratives in BookSum and Project Gutenberg. While these sources provide diverse and well-structured story materials, future work may further include a broader range of narrative genres and text types, such as contemporary fiction, children’s literature, web novels, interactive stories, or domain-specific narratives, to examine the generality of preference-aligned rewriting in more varied settings.

\section*{Ethical Considerations}

We acknowledge the ethical considerations involved in constructing STORYLENSBENCH from literary texts and reader reviews.

\textbf{Data Sources and Copyright.}
The source texts are collected from BookSum and Project Gutenberg and are used only for academic research. To respect copyright and licensing constraints, we will not release copyrighted full-text books or unrestricted reproductions of protected source materials. Released data will focus on processed benchmark instances, metadata, structured contexts, and derived annotations within the permissions of the original sources.

\textbf{Reader Privacy.}
Reader preference profiles are derived from publicly available Goodreads review content. To protect user privacy, we anonymize user-related information and avoid releasing raw user identifiers, review identifiers, or information that could directly identify individual readers. We also avoid distributing raw review texts whenever possible and release only processed preference representations needed for research.

\textbf{Intended Use.}
The benchmark, evaluator, and rewriting model are intended for academic research on preference-aligned narrative rewriting. They should not be used to impersonate specific readers, infer sensitive personal attributes, or generate misleading adaptations of copyrighted works.

\bibliography{main}

\clearpage
\appendix

\section{Dataset Details}
\label{app:dataset_analysis}
\begin{figure}[t]
    \centering
    \includegraphics[
        width=\columnwidth,
        trim=0 0 0 0,
        clip
    ]{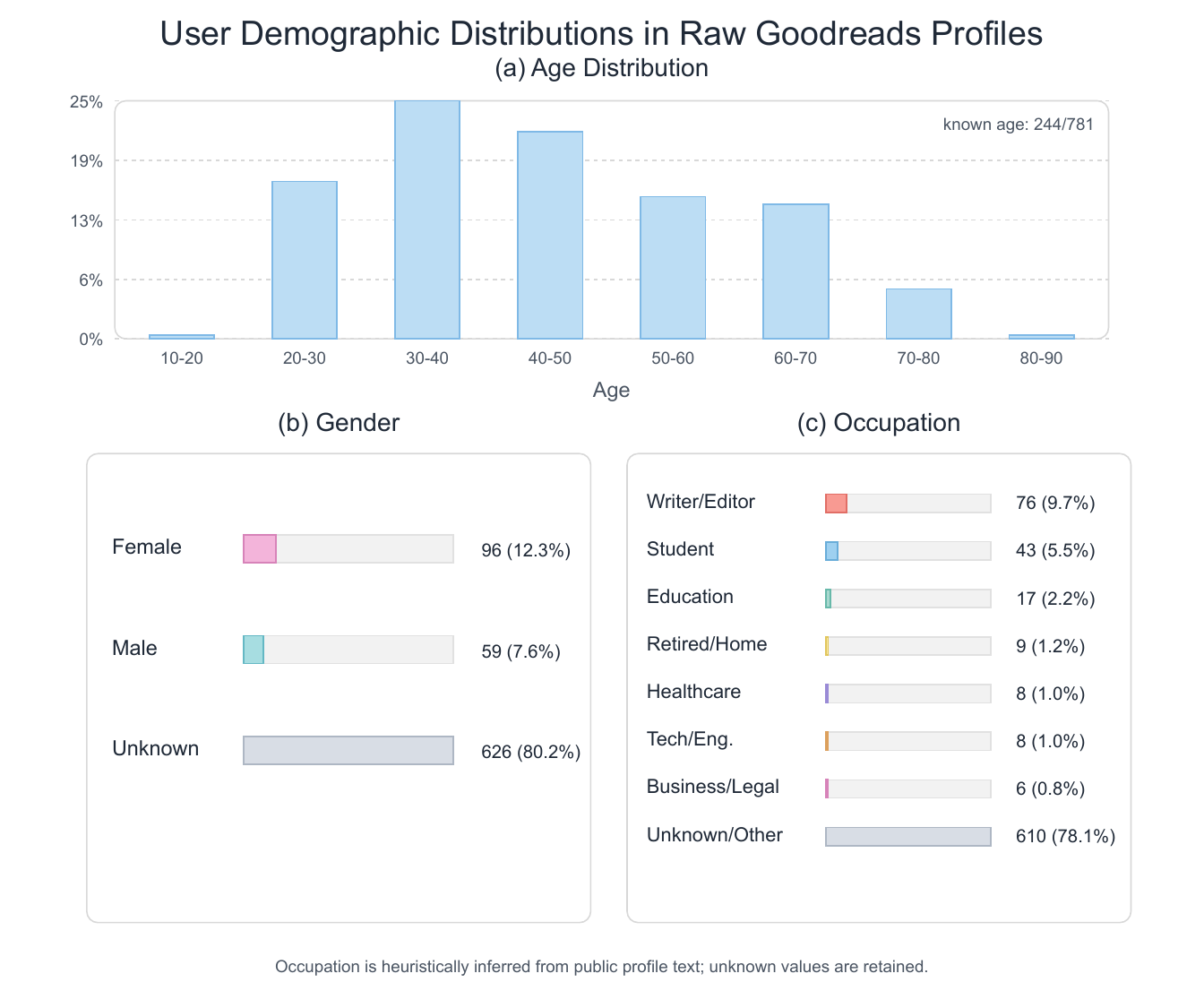}
    \caption{Demographic distributions of raw Goodreads reviewer profiles.}
    \label{fig:goodreads_demographics}
\end{figure}
\begin{figure}[t]
    \centering
    \includegraphics[
        width=\columnwidth,
        trim=20mm 0 10mm 0,
        clip
    ]{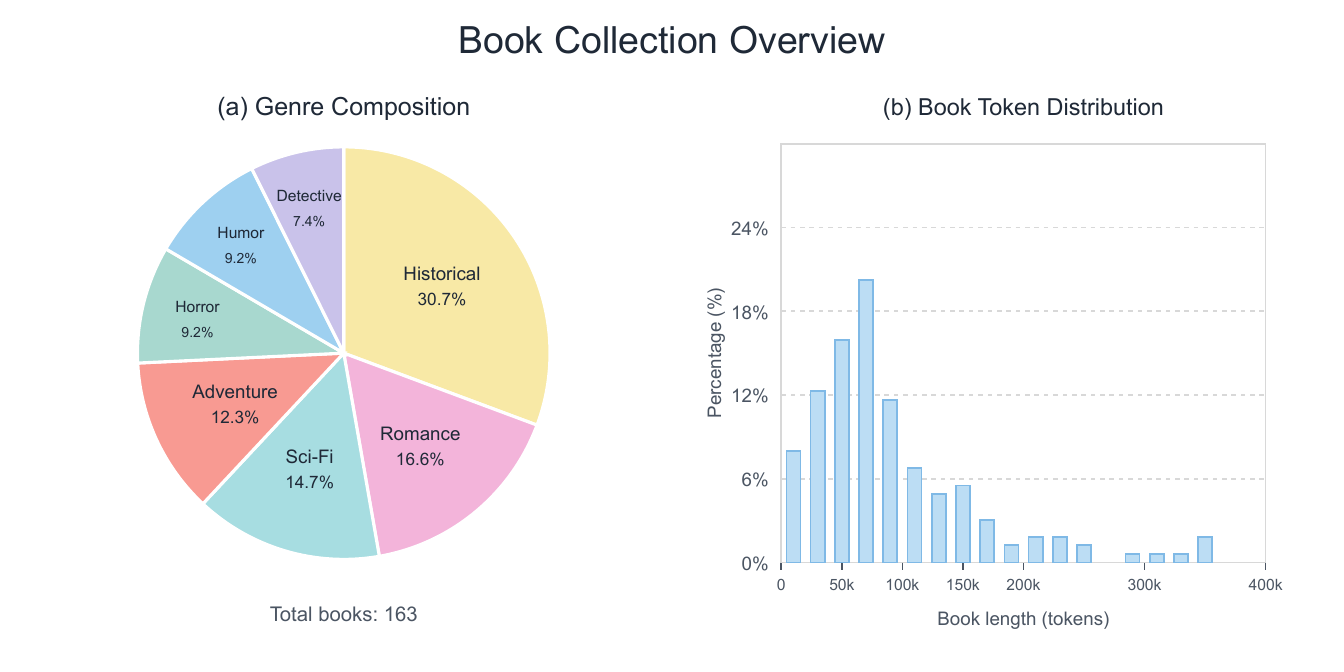}
    \caption{Book genre and length distributions in \textsc{StoryLensBench}.}
    \label{fig:book_overview}
\end{figure}
\begin{figure*}[t]
    \centering
    \includegraphics[
        width=\textwidth,
        trim=8mm 10mm 24mm 24mm,
        clip
    ]{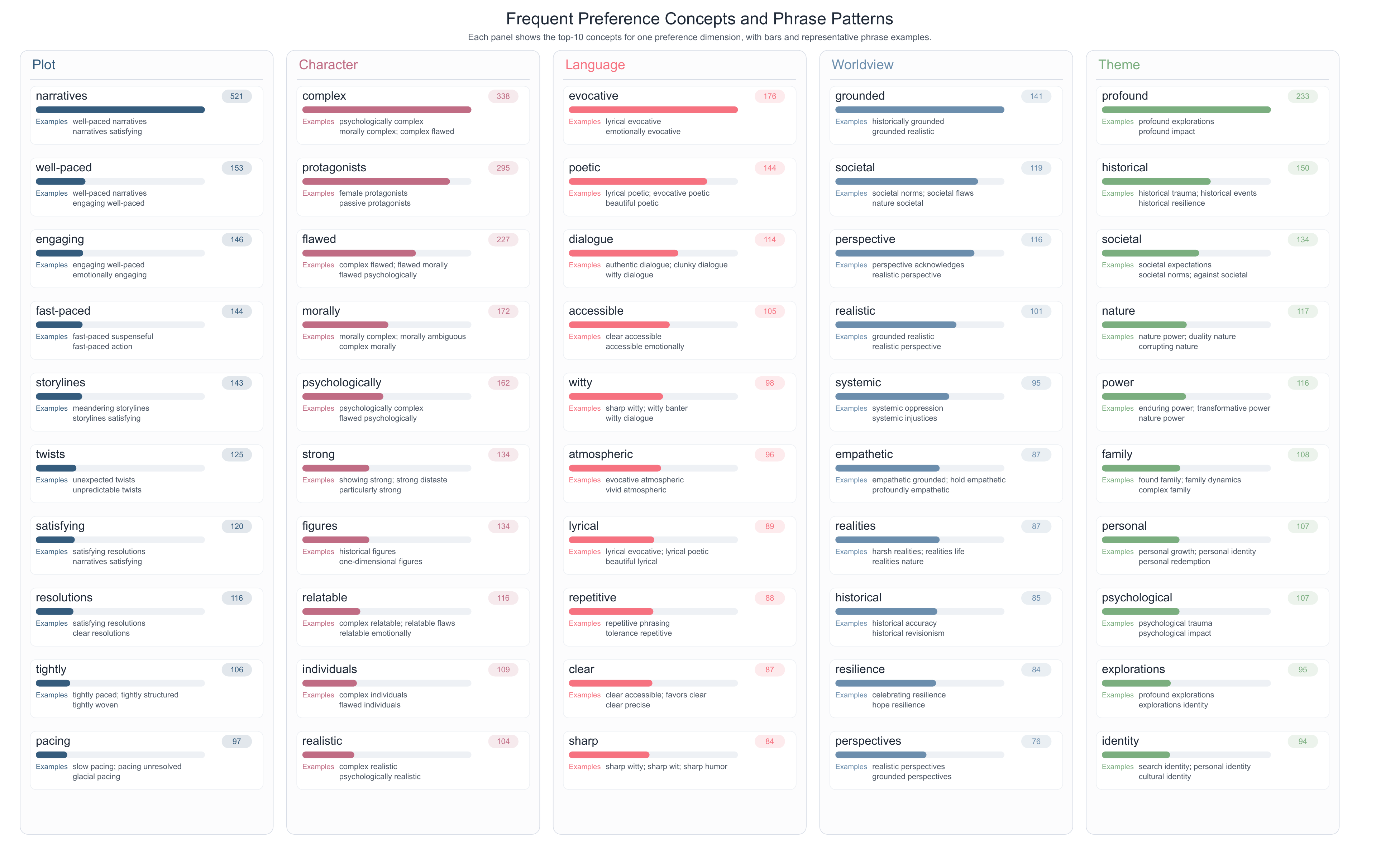}
    \caption{Top keywords in user preferences across five preference dimensions.}
    \label{fig:top_keywords_by_field}
\end{figure*}

Figure~\ref{fig:top_keywords_by_field} shows the keyword distributions of user preferences across five dimensions: plot, character, language, worldview, and theme. 
For each dimension, we report the top-10 most frequent keywords and provide representative phrases for each keyword to illustrate how these preferences are expressed in natural language. 
Across all dimensions, the distributions show a clear long-tail pattern.

This indicates that user preferences contain both common patterns and diverse personalized details. 
Therefore, preference-aligned story rewriting needs to capture not only dominant preference trends but also fine-grained individual signals.

\section{Pilot Study Results}
\label{app:pilot_study}

This section reports the detailed results of the two pilot studies mentioned in the Introduction.

\subsection{Pilot Study A}

Tables~\ref{tab:pilot_a_lexical} and~\ref{tab:pilot_a_syntax} show the lexical and syntactic adaptation results. 
Gemini 2.5 Pro~\citep{google2025gemini25pro} and GPT-5~\citep{openai2025gpt5} perform better in vocabulary control, while GPT-5 achieves the lowest syntactic complexity.
However, the differences across models, including Claude 3.5~\citep{anthropic2024claude35} and DeepSeek R1~\citep{deepseek2025r1}, indicate that even strong LLMs still vary in satisfying fine-grained audience constraints.

\subsection{Pilot Study B}

Table~\ref{tab:pilot_b_summary} summarizes the human annotation results across eight books. 
Context-enhanced rewriting achieves the highest satisfaction and overall score, while the original text preserves the highest fidelity. 
This suggests that context-consistent enrichment improves reader satisfaction, but may introduce a modest fidelity trade-off.

\begin{table}[t]
\centering
\small
\begin{tabular*}{\columnwidth}{@{\extracolsep{\fill}}lc}
\toprule
\textbf{Metric} & \textbf{Average Value} \\
\midrule
Preference alignment & 2.16 \\
Fluency & 2.36 \\
Fidelity & 2.44 \\
Ranking agreement & 81\% \\
\bottomrule
\end{tabular*}
\caption{Human verification results for sampled rewritten stories and ranking pairs. The first three dimensions are rated by annotators on a 0--3 scale, where higher scores indicate better rewriting quality. Ranking agreement measures the average human agreement with model-produced rankings.}
\label{tab:human_verification}
\end{table}
\begin{table*}[t]
\centering
\begin{minipage}[t]{0.49\textwidth}
\vspace{0pt}
\centering
\small
\resizebox{\linewidth}{!}{
\begin{tabular}{lccc}
\toprule
\textbf{Model}
& \textbf{A1+A2 Basic Words $\uparrow$}
& \textbf{B2+C1+C2 Advanced Words $\downarrow$}
& \textbf{Interpretation} \\
\midrule
Gemini 2.5 Pro
& \textbf{92.9\%}
& \textbf{2.8\%}
& Best adapted to children \\
GPT-5
& 91.6\%
& 3.6\%
& Strong lexical adaptation \\
Claude 3.5
& 89.1\%
& 4.9\%
& Slightly more difficult wording \\
DeepSeek R1
& 86.8\%
& 6.6\%
& More advanced wording \\
\bottomrule
\end{tabular}
}
\par\smallskip
\refstepcounter{table}\label{tab:pilot_a_lexical}
\raggedright
\textbf{Table \thetable}. Pilot A lexical adaptation results. We compare the proportion of basic and advanced words in rewritten stories for children.
\end{minipage}
\hfill
\begin{minipage}[t]{0.49\textwidth}
\vspace{0pt}
\centering
\small
\resizebox{\linewidth}{!}{
\begin{tabular}{lcccc}
\toprule
\textbf{Model}
& \textbf{MLS $\downarrow$}
& \textbf{C/S $\downarrow$}
& \textbf{CN/T $\downarrow$}
& \textbf{Interpretation} \\
\midrule
GPT-5
& \textbf{6.7}
& \textbf{1.18}
& \textbf{0.41}
& Best syntactic adaptation \\
Gemini 2.5 Pro
& 7.7
& 1.41
& 0.60
& Adapted to children \\
DeepSeek R1
& 11.9
& 1.21
& 1.28
& More suitable for adolescent readers \\
Claude 3.5
& 15.7
& 1.62
& 1.59
& Adult-like syntactic style \\
\bottomrule
\end{tabular}
}
\par\smallskip
\refstepcounter{table}\label{tab:pilot_a_syntax}
\raggedright
\textbf{Table \thetable}. Pilot A syntactic complexity results. MLS denotes mean length of sentence, C/S denotes clauses per sentence, and CN/T denotes complex nominals per T-unit.
\end{minipage}
\end{table*}

\begin{table*}[t]
\centering
\small
\resizebox{\textwidth}{!}{
\begin{tabular}{lccccc}
\toprule
\textbf{Method} 
& \textbf{Satisfy $\uparrow$} 
& \textbf{Fluency $\uparrow$} 
& \textbf{Fidelity $\uparrow$} 
& \textbf{Total Avg. $\uparrow$} 
& \textbf{Best Count} \\
\midrule
Original 
& 3.43 
& 4.39 
& \textbf{4.62} 
& 4.15 
& 1 \\
Preference-alignment 
& 3.51 
& 4.39 
& 4.46 
& 4.12 
& 2 \\
Context-enhanced 
& \textbf{4.27} 
& \textbf{4.45} 
& 4.41 
& \textbf{4.37} 
& \textbf{6} \\
\bottomrule
\end{tabular}
}
\caption{
Pilot B human annotation summary across eight books. Context-enhanced rewriting achieves the highest satisfaction and overall score, while the original text preserves the highest fidelity.
}
\label{tab:pilot_b_summary}
\end{table*}

\section{Case Study}
\noindent\textbf{Case 1.}
Table~\ref{tab:case1_robinson_crusoe} presents a case from \textit{Robinson Crusoe}, Chapter IV. The key gain from context enhancement is that commercial language is not added as a surface style alone; it is tied back to Crusoe's already established background as a plantation owner, his pursuit of wealth, and his methodical survival logic. As a result, phrases such as \textbf{repository of capital}, \textbf{merchant assessing an abandoned warehouse}, and \textbf{means of production} function as continuity-grounded characterization rather than decorative preference matching. The final phrase \textbf{catastrophic loss of equity} extends the same logic from objects to risk management, making the raft legible as an accumulated asset system rather than just a transport device.

\begin{table*}[t]
\centering
\scriptsize
\setlength{\tabcolsep}{4pt}
\renewcommand{\arraystretch}{1.10}
\begin{tabular}{>{\raggedright\arraybackslash}p{0.22\textwidth}>{\raggedright\arraybackslash}p{0.18\textwidth}>{\raggedright\arraybackslash}p{0.25\textwidth}>{\raggedright\arraybackslash}p{0.25\textwidth}}
\toprule
\textbf{Original Excerpt} & \textbf{Reader Preference Profile} & \textbf{Preference-alignment Rewrite (no context)} & \textbf{Context-enhanced Rewrite (ours)} \\
\midrule
Here I found that the ship was bulged, and had a great deal of water in her hold, but that she lay so on the side of a bank of hard sand, or, rather earth, that her stern lay lifted up upon the bank, and her head low, almost to the water. By this means all her quarter was free, and all that was in that part was dry; for you may be sure my first work was to search, and to see what was spoiled and what was free.\par

And, first, I found that all the ship's provisions were dry and untouched by the water, and being very well disposed to eat, I went to the bread room and filled my pockets with biscuit, and ate it as I went about other things, for I had no time to lose. I also found some rum in the great cabin, of which I took a large dram, and which I had, indeed, need enough of to spirit me for what was before me.\par

My raft was now strong enough to bear any reasonable weight. My next care was what to load it with, and how to preserve what I laid upon it from the surf of the sea. ... I had other things which my eye was more upon - as, first, tools to work with on shore. And it was after long searching that I found out the carpenter's chest, which was, indeed, a very useful prize to me, and much more valuable than a shipload of gold would have been at that time.
&
\textbf{plot:} mercantile strategy, economic negotiation, trade networks.\par
\textbf{character:} shrewd pragmatists and adaptable mediators.\par
\textbf{language:} measured prose with material culture and socio-economic detail.\par
\textbf{worldview:} resource allocation, incentives, exchange.\par
\textbf{theme:} commerce, wealth, soft power, globalization.
&
I discovered that the ship's hull was breached and its hold was flooded. Fortunately, it had run aground on a hard sandbank in such a way that the stern was lifted high, keeping that entire rear section perfectly dry. My \textbf{immediate priority} was to see what had been ruined and what was \textbf{salvageable}.\par

What I needed most was a way to transport the essential supplies back to shore. Wishing for a boat was useless, so necessity forced me into action. ... Driven by the \textbf{desperate need for supplies}, I pushed through the exhaustion, grabbed a carpenter's saw, and cut a spare topmast into three heavy pieces to reinforce my makeshift raft.\par

My mind was focused on a much higher priority: tools. After a long search, I finally located the ship's carpenter's chest. ... I lowered the heavy box onto my raft, viewing it as a prize \textbf{far more precious ... than a ship entirely filled with gold}.
&
The morning light revealed a profound shift in my material circumstances. ... I recognized an immediate imperative: the ship was no longer a vessel of transit, but a \textbf{repository of capital} waiting to be liquidated before the sea could reclaim it.\par

Stripping myself of burdensome garments, I swam to the wreck, approaching it as a \textbf{merchant assessing an abandoned warehouse}. ... The ship's quarter remained elevated and dry, preserving the vital commodities within.\par

Lacking a skiff, I engineered a rudimentary barge ... My foremost priority was the \textbf{means of production}: the ship's carpenter's chest. In an environment devoid of manufactured goods, these tools represented an incalculable concentration of industrial potential, far surpassing the value of any precious metal.\par

Navigating the laden raft to the shore required a pragmatic application of the tides. ... A grounding upon a sudden shoal threatened to capsize the venture, an event that would have constituted a \textbf{catastrophic loss of equity}.
\\
\bottomrule
\end{tabular}
\caption{Case 1 from \textit{Robinson Crusoe}. Structured context grounds mercantile preference cues in Crusoe's established history and survival logic.}
\label{tab:case1_robinson_crusoe}
\end{table*}

\noindent\textbf{Case 2.}
Tables~\ref{tab:local_fidelity_examples}--\ref{tab:fidelity_judgment_criteria} illustrate how we judge local and global fidelity with concrete examples. We first show claim-level local fidelity decisions, then a globally acceptable literary expansion that is locally unsupported, and finally summarize the decision criteria.

\begin{table*}[t]
\centering
\small
\setlength{\tabcolsep}{4pt}
\renewcommand{\arraystretch}{1.12}
\begin{tabular}{>{\raggedright\arraybackslash}p{0.10\textwidth}>{\raggedright\arraybackslash}p{0.19\textwidth}>{\raggedright\arraybackslash}p{0.19\textwidth}>{\raggedright\arraybackslash}p{0.13\textwidth}>{\raggedright\arraybackslash}p{0.24\textwidth}}
\toprule
\textbf{Example} & \textbf{Original SVO Event} & \textbf{Rewritten Event} & \textbf{Judgment} & \textbf{Reason} \\
\midrule
Local-1 & \textless Mrs.~Reed, appoint, a small closet for Jane Eyre to sleep in\textgreater{} & \textless Mrs.~Reed, banish, Jane Eyre to a closet\textgreater{} & \textbf{Pass / Entailment} & Although appoint is rewritten as the stronger verb banish, the core event is preserved: Mrs.~Reed sends Jane Eyre to a closet. The rewrite changes the emotional tone but does not alter the main factual content. \\

Local-2 & \textless Watson, receive, the wedding ring\textgreater{} & \textless Holmes, press, a near-perfect replica into Watson's palm\textgreater{} & \textbf{Fail / Contradiction} & The rewrite changes the key object from the actual wedding ring to a near-perfect replica. This is not a stylistic paraphrase, but a factual change that alters the plot semantics. \\

Local-3 & \textless Holmes, follow, the old woman\textgreater{} & \textless Watson, watch, Holmes trail the old woman\textgreater{} & \textbf{Pass / Entailment} & The core event is still preserved: Holmes follows / trails the old woman. The rewrite adds Watson as an observer, which is a perspective expansion rather than a contradiction. \\
\bottomrule
\end{tabular}
\caption{Local Fidelity Examples}
\label{tab:local_fidelity_examples}
\end{table*}

\begin{table*}[!t]
\centering
\small
\setlength{\tabcolsep}{3pt}
\renewcommand{\arraystretch}{1.12}
\begin{tabular}{%
>{\raggedright\arraybackslash}p{0.16\textwidth}
>{\raggedright\arraybackslash}p{0.17\textwidth}
>{\raggedright\arraybackslash}p{0.13\textwidth}
>{\raggedright\arraybackslash}p{0.17\textwidth}
>{\raggedright\arraybackslash}p{0.23\textwidth}}
\toprule
\textbf{Extracted Claim} & \textbf{Local Evidence} & \textbf{Initial Local Judgment} & \textbf{Global Fidelity Judgment} & \textbf{Reason} \\
\midrule
{}[State] The kitchen usually has the scent of baking bread. &
The local source text does not explicitly mention the smell of baking bread in the kitchen. &
\textbf{Unsupported} &
\textbf{Neutral / Acceptable Literary Expansion} &
The claim is not directly entailed by the local passage, but it does not contradict the global narrative world or character setting. It functions as environmental description and literary enrichment, so it should not be penalized as hallucination. \\
\bottomrule
\end{tabular}
\caption{Global Fidelity Example}
\label{tab:global_fidelity_example}
\end{table*}

\begin{table*}[t]
\centering
\scriptsize
\setlength{\tabcolsep}{3pt}
\renewcommand{\arraystretch}{1.05}
\begin{tabular}{%
>{\raggedright\arraybackslash}p{0.15\textwidth}
>{\raggedright\arraybackslash}p{0.34\textwidth}
>{\centering\arraybackslash}p{0.12\textwidth}
>{\raggedright\arraybackslash}p{0.27\textwidth}}
\toprule
\textbf{Judgment Type} & \textbf{Definition} & \textbf{Should Penalize?} & \textbf{Example} \\
\midrule
\textbf{Entailment} &
The rewritten content is directly supported by the original text, or preserves the same core event with minor stylistic changes. &
No &
Mrs.~Reed appoints a closet for Jane Eyre $\rightarrow$ Mrs.~Reed banishes Jane Eyre to a closet \\

\textbf{Neutral} &
The rewritten content is not explicitly stated in the local source, but remains globally plausible and does not violate character, plot, or world settings. &
No &
Adding the scent of baking bread to a kitchen scene \\

\textbf{Contradiction} &
The rewritten content changes key plot facts, object identity, causal relations, or violates global character/world constraints. &
Yes &
Watson receives the wedding ring $\rightarrow$ Holmes gives Watson a replica \\
\bottomrule
\end{tabular}
\caption{Fidelity Judgment Criteria}
\label{tab:fidelity_judgment_criteria}
\end{table*}

\section{Prompt Templates}
\label{app:prompt_templates}

Appendix~D summarizes the prompt templates used in our experiments. Table~\ref{tab:training_prompt_templates} reports our training prompts, including the SFT prompt and the GRPO prompt. Table~\ref{tab:baseline_prompt_templates} reports our baseline prompts, including the direct LLM baseline prompt and the Rewrite Few-shot RAG prompt.

\begin{table*}[t]
\centering
\scriptsize
\setlength{\tabcolsep}{4pt}
\renewcommand{\arraystretch}{1.12}
\begin{tabular}{>{\raggedright\arraybackslash}p{0.12\textwidth}>{\raggedright\arraybackslash}p{0.24\textwidth}>{\raggedright\arraybackslash}p{0.56\textwidth}}
\toprule
\textbf{Method} & \textbf{Prompt Part} & \textbf{Template / Content} \\
\midrule
\multirow{3}{*}{SFT}
& Shared system prompt
& \texttt{You are a literary rewriting assistant.} The prompt instructs the model to rewrite the source chapter for the target reader while preserving plot events, causal structure, chapter continuity, character consistency, and English-only output. It also states that continuity overrides preferences and clarity overrides stylistic excess. \\

& User prompt template
& \texttt{Rewrite the source chapter for the target reader.} \texttt{Metadata: \{compact\_json\_header\}} \texttt{Structured context: \{compact\_json\_context\}} \texttt{Reader preference profile: \{compact\_json\_preference\}} \texttt{Source chapter: \{source\_text\}} \texttt{Return only the rewritten chapter text.} \\

& Assistant target
& \texttt{\{rewritten\_chapter\_text\}} \\
\midrule
\multirow{3}{*}{GRPO}
& System prompt
& The GRPO prompt inherits the SFT system prompt and adds two extra constraints: \texttt{Do not introduce new major characters, locations, events, or facts unless they are strongly implied by the source chapter and continuity context.} It also requires the rewrite to remain \texttt{roughly similar in length to the source chapter}. \\

& User prompt
& Same user prompt template as SFT, using \texttt{\{compact\_json\_header\}}, \texttt{\{compact\_json\_context\}}, \texttt{\{compact\_json\_preference\}}, and \texttt{\{source\_text\}} as input fields. \\

& Training record shape
& Each GRPO example stores \texttt{messages} with a system prompt and a user prompt, together with auxiliary fields including \texttt{preference\_text}, \texttt{source\_text}, \texttt{reference\_rewrite}, chapter metadata, file paths, and a nested \texttt{metadata} object. \\
\bottomrule
\end{tabular}
\caption{Training prompt templates used for SFT and GRPO.}
\label{tab:training_prompt_templates}
\end{table*}

\begin{table*}[t]
\centering
\scriptsize
\setlength{\tabcolsep}{4pt}
\renewcommand{\arraystretch}{1.12}
\begin{tabular}{>{\raggedright\arraybackslash}p{0.14\textwidth}>{\raggedright\arraybackslash}p{0.22\textwidth}>{\raggedright\arraybackslash}p{0.56\textwidth}}
\toprule
\textbf{Method} & \textbf{Prompt Part} & \textbf{Template / Content} \\
\midrule
\multirow{2}{*}{LLM Baseline}
& System prompt
& The baseline uses the same literary rewriting system prompt as SFT: rewrite for the target reader, preserve main plot events and causal structure, maintain cross-chapter continuity, keep characters consistent, adapt to the five preference dimensions, output only story text, and keep the result in English. \\

& User prompt template
& \texttt{Rewrite the source chapter for the target reader.} \texttt{Metadata: \{compact\_json\_header\}} \texttt{Structured context: \{compact\_json\_context\}} \texttt{Reader preference profile: \{compact\_json\_preference\}} \texttt{Source chapter: \{source\_text\}} \texttt{Return only the rewritten chapter text.} \\
\midrule
\multirow{3}{*}{Rewrite Few-shot RAG}
& Retrieval query
& \texttt{\{target\_profile\} \{target\_original\}} \\

& Final prompt template
& The final prompt repeats the same rewriting instruction block and then appends a retrieved few-shot section: \texttt{--- Reference Examples (Combine these styles) ---}. Each example contains \texttt{[Instruction]}, \texttt{[Original]}, and \texttt{[Rewrite]} fields, followed by the current task with metadata, structured context, reader preference profile, and source chapter. \\

& Example record schema
& Retrieved demonstrations are formatted as \texttt{[Instruction]: data.json -> instruction}, \texttt{[Original]: data.json -> original\_text}, and \texttt{[Rewrite]: data.json -> rewritten\_text}. \\
\bottomrule
\end{tabular}
\caption{Baseline prompt templates used for direct LLM baselines and Rewrite Few-shot RAG.}
\label{tab:baseline_prompt_templates}
\end{table*}

\section{Implementation Details}
\label{app:implementation_details}

We train \textsc{StoryLensWriter} in two stages: supervised fine-tuning (SFT) followed by GRPO. The SFT stage is implemented with ms-swift and DeepSpeed ZeRO-3 on 8 Nvidia A6000 GPUs. We use LoRA with rank 8 and alpha 16, and the full SFT run takes about 36.4 hours.

We then perform GRPO training with VeRL on 8 Nvidia A6000 GPUs. In this stage, 5 GPUs are allocated to actor/reference training, 2 GPUs to vLLM rollout generation, and 1 GPU to reward computation. We use LoRA with rank 8 and alpha 32, and the GRPO run takes about 49.6 hours. All final results reported in the paper use the same decode configuration: \texttt{max\_new\_tokens}=3072, temperature 0.25, top-$p$ 0.8, top-$k$ 20, and repetition penalty 1.05. We use this setting as a practical trade-off between creative variation and overall text quality.




\end{document}